\documentclass[conference]{IEEEtran}

\IEEEoverridecommandlockouts
\usepackage{cite}
\usepackage{amsmath,amssymb,amsfonts}
\usepackage{algorithmic}
\usepackage{graphicx}
\usepackage{textcomp}
\usepackage{xcolor}
\usepackage{multirow}
\usepackage[squaren, Gray, cdot]{SIunits}
\usepackage{balance}

\def\BibTeX{{\rm B\kern-.05em{\sc i\kern-.025em b}\kern-.08em
    T\kern-.1667em\lower.7ex\hbox{E}\kern-.125emX}}

\usepackage[lofdepth,lotdepth]{subfig}
\usepackage{tikz}
\usepackage{pgfplots}
\usepackage{filecontents}
\usepackage[skins]{tcolorbox}
\usepackage[toc, acronym]{glossaries}
\usepackage{booktabs}
\usepackage{hyperref}

\usetikzlibrary{positioning, fit, calc, arrows.meta}
\usepgfplotslibrary{statistics}

\definecolor{plotcolor1}{RGB}{57,106,177}
\definecolor{plotcolor2}{RGB}{204,37,41}
\definecolor{plotcolor3}{RGB}{62,150,81}
\definecolor{plotcolor4}{RGB}{218,124,48}
\definecolor{plotcolor5}{RGB}{83,81,84}
\definecolor{plotcolor6}{RGB}{107,76,154}
\definecolor{plotcolor7}{RGB}{146,36,40}
\definecolor{plotcolor8}{RGB}{148,139,61}

\definecolor{histocolor1}{RGB}{114,147,203}
\definecolor{histocolor2}{RGB}{211,94,96}
\definecolor{histocolor3}{RGB}{132,186,91}
\definecolor{histocolor4}{RGB}{225,151,76}
\definecolor{histocolor5}{RGB}{128,133,133}
\definecolor{histocolor6}{RGB}{144,103,167}
\definecolor{histocolor7}{RGB}{171,104,87}
\definecolor{histocolor8}{RGB}{204,194,16}

\newacronym{ann}{ANN}{artificial neural network}
\newacronym{dnn}{DNN}{deep neural network}
\newacronym{snn}{SNN}{spiking neural network}
\newacronym{cnn}{CNN}{Convolutional neural network}
\newacronym{stdp}{STDP}{spike-timing-dependent plasticity}
\newacronym{ltd}{LTD}{long-term depression}
\newacronym{ltp}{LTP}{long-term potentiation}
\newacronym{n2s3}{N2S3}{neural network scalable spiking simulator}
\newacronym{fpga}{FPGA}{field-programmable gate array}
\newacronym{fppa}{FPPA}{field-programmable analog array}
\newacronym{vlsi}{VLSI}{very large-scale integration}
\newacronym{cmos}{CMOS}{complementary metal-oxide semiconductor}
\newacronym{rgb}{RGB}{red green blue}
\newacronym{rbm}{RBM}{restricted Boltzmann machine}
\newacronym{cd}{CD}{contrastive divergence}
\newacronym{dbn}{DBN}{deep belief network}
\newacronym{gd}{GD}{gradient descent}
\newacronym{sgd}{SGD}{stochastic gradient descent}
\newacronym{bp}{BP}{back-propagation}
\newacronym{mnist}{MNIST}{Modified-NIST}
\newacronym{tpu}{TPU}{tensor processor unit}
\newacronym{if}{IF}{integrate-and-fire}
\newacronym{lif}{LIF}{leaky integrate-and-fire}
\newacronym{eif}{EIF}{exponential integrate-and-fire}
\newacronym{qif}{QIF}{quadratic integrate-and-fire}
\newacronym{adex}{LIF}{adaptive exponential integrate-and-fire}
\newacronym{lat}{LAT}{leaky adaptive threshold}
\newacronym{tft}{TFT}{target frequency threshold}
\newacronym{wta}{WTA}{winner-takes-all}
\newacronym{lstm}{LSTM}{long short-term memory}
\newacronym{mlp}{MLP}{multi-layer perceptron}
\newacronym{ae}{AE}{auto-encoder}
\newacronym{dog}{DoG}{difference of Gaussians}
\newacronym{relu}{ReLU}{rectified linear unit}
\newacronym{rbf}{RBF}{radial basis function}
\newacronym{svm}{SVM}{support vector machine}
\newacronym{dsl}{DSL}{domain specific language}
\newacronym{rc}{RC}{reservoir computing}
\newacronym{ff}{FF}{feed-forward}
\newacronym{csnns}{CSNNS}{convolutional spiking neural network simulator}
\newacronym{simd}{SIMD}{single instruction multiple data}
\newacronym{mimd}{MIMD}{multiple instruction multiple data}
\newacronym{gpu}{GPU}{graphical processing unit}
\newacronym{cpu}{CPU}{central processing unit}
\newacronym{mse}{MSE}{mean squared error}
\newacronym{sift}{SIFT}{scale-invariant feature transform}
\newacronym{iot}{IoT}{Internet of things}
\newacronym{psp}{PSP}{post-synaptic potential}
\newacronym{zca}{ZCA}{zero component analysis}
\newacronym{pca}{PCA}{principal component analysis}
\newacronym{url}{URL}{uniform resource locator}
\newacronym{reram}{ReRAM}{resistive RAM}
\newacronym{pcm}{PCM}{phase change memory}
\newacronym{sttram}{STT-RAM}{spin-torque transfer RAM}
\newacronym{nomfet}{NOMFET}{nanoparticle organic memory field-effect transistor}
\newacronym{ea}{EA}{evolutionary algorithm}
\newacronym{dvs}{DVS}{dynamic vision sensor}
\newacronym{jvm}{JVM}{Java virtual machine}
\newacronym{aer}{AER}{address-event representation}
\newacronym{surf}{SURF}{speeded-up robust features descriptor}
\newacronym{bcm}{BCM}{Bienenstock-Cooper-Munro}

\newglossaryentry{m_dirac}{name=\ensuremath{\delta}, description={Dirac function}}
\newglossaryentry{m_time}{name=\ensuremath{t}, description={Timestamp}}
\newglossaryentry{m_learning_rate}{name=\ensuremath{\eta}, description={Learning rate}}
\newglossaryentry{m_gamma}{name=\ensuremath{\gamma}, description={Update factor}}
\newglossaryentry{m_convolution}{name=\ensuremath{\ast}, description={Convolution operator}}
\newglossaryentry{m_kullback_liebler}{name=\ensuremath{\operatorname{KL}}, description={Kullback-Liebler divergence}}
\newglossaryentry{m_sparseness}{name=\ensuremath{\operatorname{sp}}, description={Sparseness measure}}
\newglossaryentry{m_coherence}{name=\ensuremath{\mu}, description={Coherence measure}}
\newglossaryentry{m_input_value}{name=\ensuremath{x}, description={Input value}}
\newglossaryentry{m_output_value}{name=\ensuremath{y}, description={Output value}}
\newglossaryentry{m_epoch_number}{name=\ensuremath{n_{\operatorname{epoch}}}, description={Number of epochs}}
\newglossaryentry{m_annealing}{name=\ensuremath{\alpha}, description={Annealing factor}}
\newglossaryentry{m_identity_matrix}{name=\ensuremath{\operatorname{I}}, description={Identity matrix}}
\newglossaryentry{m_normal_distribution}{name=\ensuremath{\mathcal{G}}, description={Normal distribution}}
\newglossaryentry{m_uniform_distribution}{name=\ensuremath{\mathcal{U}}, description={Uniform distribution}}
\newglossaryentry{m_perpendicular_distance}{name=\ensuremath{\operatorname{pd}}, description={Perpendicular distance}}
\newglossaryentry{m_cov_matrix}{name=\ensuremath{\operatorname{\Sigma}}, description={Covariance matrix}}
\newglossaryentry{m_eigen_ratio}{name=\ensuremath{\rho}, description={Ratio of non-null eigenvector}}
\newglossaryentry{m_eigen_vector_matrix}{name=\ensuremath{\operatorname{U}}, description={Eigenvectors matrix}}
\newglossaryentry{m_eigen_value}{name=\ensuremath{\lambda}, description={Eigenvalue}}
\newglossaryentry{m_eigen_value_matrix}{name=\ensuremath{\operatorname{\Lambda}}, description={Eigenvalue diagonal matrix}}

\newglossaryentry{m_max_perpendicular_distance}{name=\ensuremath{\operatorname{pd}_{\max}}, description={Perpendicular distance}}
\newglossaryentry{m_trajectory_orientation}{name=\ensuremath{m_{\Theta}}, description={Trajectory orientation}}
\newglossaryentry{m_trajectory_min_orientation}{name=\ensuremath{m_{\min\Theta}}, description={Minimum trajectory orientation}}
\newglossaryentry{m_trajectory_max_orientation}{name=\ensuremath{m_{\max\Theta}}, description={Maximum trajectory orientation}}
\newglossaryentry{m_trajectory_shift}{name=\ensuremath{m_{\operatorname{S}}}, description={Trajectory shift}}
\newglossaryentry{m_trajectory_velocity}{name=\ensuremath{m_{\operatorname{V}}}, description={Trajectory velocity}}

\newglossaryentry{m_image}{name=\ensuremath{\operatorname{X}}, description={Image matrix}}
\newglossaryentry{m_image_reconstruction}{name=\ensuremath{\operatorname{\hat{X}}}, description={Image reconstruction}}
\newglossaryentry{m_image_width}{name=\ensuremath{x_{\operatorname{width}}}, description={Image width}}
\newglossaryentry{m_image_height}{name=\ensuremath{x_{\operatorname{height}}}, description={Image height}}
\newglossaryentry{m_image_depth}{name=\ensuremath{x_{\operatorname{depth}}}, description={Image depth}}
\newglossaryentry{m_pulse_martix}{name=\ensuremath{\operatorname{J}}, description={Impulse response matrix}}

\newglossaryentry{m_feature}{name=\ensuremath{g}, description={Feature}}
\newglossaryentry{m_feature_vector}{name=\ensuremath{\operatorname{g}}, description={Feature vector}}
\newglossaryentry{m_feature_number}{name=\ensuremath{n_{\operatorname{features}}}, description={Number of features}}

\newglossaryentry{m_class}{name=\ensuremath{c}, description={Class label}}
\newglossaryentry{m_class_set}{name=\ensuremath{\mathcal{C}}, description={Class set}}

\newglossaryentry{m_image_train_set}{name=\ensuremath{\mathcal{X}_{\operatorname{train}}}, description={Image train set}}
\newglossaryentry{m_label_train_set}{name=\ensuremath{\mathcal{Y}_{\operatorname{train}}}, description={Label train set}}
\newglossaryentry{m_image_test_set}{name=\ensuremath{\mathcal{X}_{\operatorname{test}}}, description={Image test set}}
\newglossaryentry{m_label_test_set}{name=\ensuremath{\mathcal{Y}_{\operatorname{test}}}, description={Label test set}}

\newglossaryentry{m_recognition_rate}{name=\ensuremath{\operatorname{rr}}, description={Recognition rate}}
\newglossaryentry{m_svm_cost}{name=\ensuremath{\operatorname{svm}_c}, description={SVM cost parameter}}
\newglossaryentry{m_hyper_parameter}{name=\ensuremath{\Phi}, description={Model hyper-parameters}}

\newglossaryentry{m_function_image_classification}{name=\ensuremath{f_{\operatorname{ec}}}, description={Image classification function}}
\newglossaryentry{m_function_feature_extractor}{name=\ensuremath{f_{\operatorname{e}}}, description={Feature extraction function}}
\newglossaryentry{m_function_classification}{name=\ensuremath{f_{\operatorname{c}}}, description={Classification function}}
\newglossaryentry{m_function_objective}{name=\ensuremath{f_{\operatorname{obj}}}, description={Objective function}}
\newglossaryentry{m_function_encoder}{name=\ensuremath{f_{\operatorname{enc}}}, description={Encoder function}}
\newglossaryentry{m_function_decoder}{name=\ensuremath{f_{\operatorname{dec}}}, description={Decoder function}}

\newglossaryentry{m_function_activation}{name=\ensuremath{f_{\sigma}}, description={Activation function}}
\newglossaryentry{m_bias}{name=\ensuremath{\operatorname{b}}, description={Bias}}

\newglossaryentry{m_norm_factor}{name=\ensuremath{\kappa}, description={Normalization factor}}
\newglossaryentry{m_sparsity_factor}{name=\ensuremath{\upsilon}, description={Sparsity factor}}

\newglossaryentry{m_expected_sparsity}{name=\ensuremath{\rho}, description={Expected sparsity}}
\newglossaryentry{m_actual_sparsity}{name=\ensuremath{\hat{\rho}}, description={Actual sparsity}}

\newglossaryentry{m_neuron_potential}{name=\ensuremath{v}, description={Membrane potential}}
\newglossaryentry{m_input_current}{name=\ensuremath{z}, description={Input current}}
\newglossaryentry{m_resting_potential}{name=\ensuremath{v_{\operatorname{rest}}}, description={Membrane resting potential}}
\newglossaryentry{m_spike_voltage}{name=\ensuremath{v_{\operatorname{exc}}}, description={Spike voltage}}
\newglossaryentry{m_inh_voltage}{name=\ensuremath{v_{\operatorname{inh}}}, description={Inhibitory spike voltage}}
\newglossaryentry{m_refractory_duration}{name=\ensuremath{t_{\operatorname{ref}}}, description={Refractory duration}}
\newglossaryentry{m_potential_leak}{name=\ensuremath{\tau_{\operatorname{leak}}}, description={Neuron leakage factor}}
\newglossaryentry{m_membrane_resistance}{name=\ensuremath{r_{\operatorname{m}}}, description={Membrane resistance}}
\newglossaryentry{m_membrane_capacitance}{name=\ensuremath{c_{\operatorname{m}}}, description={Membrane capacitance}}

\newglossaryentry{m_potential_threshold}{name=\ensuremath{v_{\operatorname{th}}}, description={Membrane potential threshold}}
\newglossaryentry{m_min_threshold}{name=\ensuremath{\operatorname{th}_{\operatorname{min}}}, description={Minimum threshold value}}
\newglossaryentry{m_threshold_lr}{name=\ensuremath{\eta_{\operatorname{th}}}, description={Threshold learning rate}}
\newglossaryentry{m_threshold_change}{name=\ensuremath{\Delta_{\operatorname{th}}}, description={Threshold update variation}}
\newglossaryentry{m_adaptive_threshold}{name=\ensuremath{\Theta}, description={Adaptive threshold}}
\newglossaryentry{m_adaptive_threshold_add}{name=\ensuremath{\Theta_{\operatorname{+}}}, description={Adaptive threshold increase constant}}
\newglossaryentry{m_adaptive_threshold_leak}{name=\ensuremath{\Theta_{\operatorname{leak}}}, description={Adaptive threshold leak time constant}}

\newglossaryentry{m_pre_frequency}{name=\ensuremath{F_{\operatorname{pre}}}, description={Pre-synaptic frequency}}
\newglossaryentry{m_post_frequency}{name=\ensuremath{F_{\operatorname{post}}}, description={Post-synaptic frequency}}
\newglossaryentry{m_threshold_frequency}{name=\ensuremath{\theta_{\operatorname{th}}}, description={Frequency threshold}}

\newglossaryentry{m_expected_frequency}{name=\ensuremath{F_{\operatorname{exp}}}, description={Expected frequency}}
\newglossaryentry{m_actual_frequency}{name=\ensuremath{F_{\operatorname{act}}}, description={Actual frequency}}
\newglossaryentry{m_frequency_difference}{name=\ensuremath{\Delta_F}, description={Relative frequency difference}}

\newglossaryentry{m_time_pre}{name=\ensuremath{t_{\operatorname{pre}}}, description={Pre-synaptic timestamp}}
\newglossaryentry{m_time_post}{name=\ensuremath{t_{\operatorname{post}}}, description={Post-synaptic timestamp}}
\newglossaryentry{m_expected_time}{name=\ensuremath{t_{\operatorname{exp}}}, description={Expected timestamp}}
\newglossaryentry{m_actual_time}{name=\ensuremath{t_{\operatorname{act}}}, description={Expected timestamp}}
\newglossaryentry{m_time_difference}{name=\ensuremath{\Delta_t}, description={Timing difference}}
\newglossaryentry{m_spike_kernel}{name=\ensuremath{f_{\operatorname{spike}}}, description={Spike kernel}}
\newglossaryentry{m_spike}{name=\ensuremath{e}, description={Spike}}
\newglossaryentry{m_spike_set}{name=\ensuremath{\mathcal{E}}, description={Spike set}}
\newglossaryentry{m_fire_set}{name=\ensuremath{\mathcal{D}}, description={Fire set}}

\newglossaryentry{m_delay}{name=\ensuremath{d}, description={Synaptic delay}}
\newglossaryentry{m_delay_change}{name=\ensuremath{\Delta_{\operatorname{d}}}, description={Synaptic delay update}}
\newglossaryentry{m_min_delay}{name=\ensuremath{d_{\operatorname{min}}}, description={Minimum synaptic delay}}
\newglossaryentry{m_max_delay}{name=\ensuremath{d_{\operatorname{max}}}, description={Maximum synaptic delay}}

\newglossaryentry{m_weight}{name=\ensuremath{w}, description={Synaptic weight}}
\newglossaryentry{m_weight_change}{name=\ensuremath{\Delta_{\operatorname{w}}}, description={Synaptic weight update}}
\newglossaryentry{m_min_weight}{name=\ensuremath{w_{\operatorname{min}}}, description={Minimum synaptic weight}}
\newglossaryentry{m_max_weight}{name=\ensuremath{w_{\operatorname{max}}}, description={Maximum synaptic weight}}
\newglossaryentry{m_weight_lr}{name=\ensuremath{\eta_{\operatorname{w}}}, description={Weight learning rate}}
\newglossaryentry{m_weight_pos_lr}{name=\ensuremath{\eta_{\operatorname{w}+}}, description={Weight learning rate in LTP}}
\newglossaryentry{m_weight_neg_lr}{name=\ensuremath{\eta_{\operatorname{w}-}}, description={Weight learning rate in LTD}}
\newglossaryentry{m_tau_stdp}{name=\ensuremath{\tau_{\operatorname{STDP}}}, description={STDP time constant}}
\newglossaryentry{m_ltp_window}{name=\ensuremath{t_{\operatorname{LTP}}}, description={LTP window duration}}
\newglossaryentry{m_update_duration}{name=\ensuremath{t_{\operatorname{update}}}, description={Update interval duration}}
\newglossaryentry{m_stdp_beta}{name=\ensuremath{\beta}, description={STDP parameter that controls the saturation effect}}

\newglossaryentry{m_actual_trace}{name=\ensuremath{r_{\operatorname{actual}}}, description={Actual synaptic activity trace}}
\newglossaryentry{m_expected_trace}{name=\ensuremath{r_{\operatorname{expected}}}, description={Expected synaptic activity trace}}

\newglossaryentry{m_neuron}{name=\ensuremath{n}, description={Neuron}}
\newglossaryentry{m_neuron_set}{name=\ensuremath{\mathcal{N}}, description={Neuron set}}
\newglossaryentry{m_synapse}{name=\ensuremath{s}, description={Synapse}}
\newglossaryentry{m_synapse_set}{name=\ensuremath{\mathcal{S}}, description={Synapse set}}
\newglossaryentry{m_layer}{name=\ensuremath{l}, description={Layer}}
\newglossaryentry{m_layer_output}{name=\ensuremath{l_{\operatorname{output}}}, description={Output layer}}
\newglossaryentry{m_layer_set}{name=\ensuremath{\mathcal{L}}, description={Layer set}}
\newglossaryentry{m_layer_width}{name=\ensuremath{l_{\operatorname{width}}}, description={Layer width}}
\newglossaryentry{m_layer_height}{name=\ensuremath{l_{\operatorname{height}}}, description={Layer height}}
\newglossaryentry{m_layer_depth}{name=\ensuremath{l_{\operatorname{depth}}}, description={Layer depth}}

\newglossaryentry{m_filter}{name=\ensuremath{h}, description={Filter}}
\newglossaryentry{m_filter_set}{name=\ensuremath{\mathcal{F}}, description={Filter set}}
\newglossaryentry{m_filter_width}{name=\ensuremath{h_{\operatorname{width}}}, description={Filter width}}
\newglossaryentry{m_filter_height}{name=\ensuremath{h_{\operatorname{height}}}, description={Filter height}}
\newglossaryentry{m_padding}{name=\ensuremath{l_{\operatorname{pad}}}, description={Padding}}
\newglossaryentry{m_stride}{name=\ensuremath{l_{\operatorname{stride}}}, description={Stride}}
\newglossaryentry{m_column}{name=\ensuremath{q}, description={Convolution column}}

\newglossaryentry{m_patch_width}{name=\ensuremath{p_{\operatorname{width}}}, description={Patch width}}
\newglossaryentry{m_patch_height}{name=\ensuremath{p_{\operatorname{height}}}, description={Patch height}}
\newglossaryentry{m_patch_number}{name=\ensuremath{n_{\operatorname{patches}}}, description={Number of patches}}
\newglossaryentry{m_pool_width}{name=\ensuremath{r_{\operatorname{width}}}, description={Pooling x size}}
\newglossaryentry{m_pool_height}{name=\ensuremath{r_{\operatorname{height}}}, description={Pooling y size}}

\newglossaryentry{m_output_width}{name=\ensuremath{o_{\operatorname{width}}}, description={Output width}}
\newglossaryentry{m_output_height}{name=\ensuremath{o_{\operatorname{height}}}, description={Output height}}
\newglossaryentry{m_output_depth}{name=\ensuremath{o_{\operatorname{depth}}}, description={Output depth}}

\newglossaryentry{m_group}{name=\ensuremath{a}, description={Sub-network}}

\newglossaryentry{m_exposition_duration}{name=\ensuremath{t_{\operatorname{exposition}}}, description={Exposition duration}}
\newglossaryentry{m_pause_duration}{name=\ensuremath{t_{\operatorname{pause}}}, description={Pause duration}}
\newglossaryentry{m_max_frequency}{name=\ensuremath{F_{\operatorname{max}}}, description={Maximum frequency}}
\newglossaryentry{m_wave_duration}{name=\ensuremath{t_{\operatorname{wave}}}, description={Wave duration}}
\newglossaryentry{m_wave_time}{name=\ensuremath{t_{\operatorname{wave}}}, description={Wave timing}}
\newglossaryentry{m_wave_deviation}{name=\ensuremath{\sigma_{\operatorname{wave}}}, description={Wave timing variance}}
\newglossaryentry{m_wave_number}{name=\ensuremath{n_{\operatorname{wave}}}, description={Number of waves generated for each sample}}
\newglossaryentry{m_input_threshold}{name=\ensuremath{x_{\operatorname{th}}}, description={Input value threshold}}
\newglossaryentry{m_function_spike_code}{name=\ensuremath{f_{\operatorname{in}}}, description={Value-to-spike conversion function}}
\newglossaryentry{m_function_spike_code_inv}{name=\ensuremath{f_{\operatorname{out}}}, description={Spike-to-value conversion function}}

\newglossaryentry{m_dog}{name=\ensuremath{\operatorname{DoG}}, description={Difference of Gaussians operator}}
\newglossaryentry{m_dog_size}{name=\ensuremath{\operatorname{DoG}_{\operatorname{size}}}, description={Size of DoG filter}}
\newglossaryentry{m_dog_center}{name=\ensuremath{\operatorname{DoG}_{\operatorname{center}}}, description={Variance of center Gaussian}}
\newglossaryentry{m_dog_surround}{name=\ensuremath{\operatorname{DoG}_{\operatorname{surround}}}, description={Variance of surround Gaussian}}
\newglossaryentry{m_pos_value}{name=\ensuremath{X_{\operatorname{+}}}, description={On channel value}}
\newglossaryentry{m_neg_value}{name=\ensuremath{X_{\operatorname{-}}}, description={Off channel value}}

\newglossaryentry{m_image_whiten}{name=\ensuremath{\operatorname{X_{\operatorname{whiten}}}}, description={Whitened image matrix}}
\newglossaryentry{m_whiten_matrix}{name=\ensuremath{\operatorname{W_{\operatorname{whiten}}}}, description={Whitening matrix}}
\newglossaryentry{m_whiten_coef}{name=\ensuremath{\epsilon}, description={Whitening coefficient}}

\newglossaryentry{m_dynamic_energy}{name=\ensuremath{e_{\operatorname{dynamic}}}, description={Dynamic energy}}
\newglossaryentry{m_static_energy}{name=\ensuremath{e_{\operatorname{static}}}, description={Static energy}}
\newglossaryentry{m_total_energy}{name=\ensuremath{e_{\operatorname{total}}}, description={Total energy}}
\newglossaryentry{m_fire_energy}{name=\ensuremath{e_{\operatorname{fire}}}, description={Energy consumed by a firing event}}
\newglossaryentry{m_spike_energy}{name=\ensuremath{e_{\operatorname{spike}}}, description={Energy consumed by a spike}}
\newglossaryentry{m_neuron_power}{name=\ensuremath{p_{\operatorname{neuron}}}, description={Power dissipated by a neuron}}
\newglossaryentry{m_synapse_power}{name=\ensuremath{p_{\operatorname{synapse}}}, description={Power dissipated by a synapse}}

\glsunset{rgb}

\begin{document}

\title{Improving STDP-based Visual Feature Learning with Whitening
\thanks{This work has been partly funded by IRCICA (Univ. Lille, CNRS,
USR 3380 – IRCICA, F-59000 Lille, France) under the Bioinspired Project.}
}

\DeclareRobustCommand*{\IEEEauthorrefmark}[1]{%
  \raisebox{0pt}[0pt][0pt]{\textsuperscript{\footnotesize #1}}%
}

\author{
\IEEEauthorblockN{
Pierre Falez\IEEEauthorrefmark{1},
Pierre Tirilly\IEEEauthorrefmark{1}\,\IEEEauthorrefmark{2},
and Ioan Marius Bilasco\IEEEauthorrefmark{1}
}
\IEEEauthorblockA{\IEEEauthorrefmark{1}
\textit{Univ. Lille, CNRS, Centrale Lille,}
\textit{UMR 9189 -- CRIStAL -- Centre de Recherche en Informatique, Signal et Automatique de Lille}\\
F-59000, Lille, France\\
}
\IEEEauthorblockA{\IEEEauthorrefmark{2}
\textit{IMT Lille Douai, F-59000, Lille, France}
}
\IEEEauthorblockA{
Email: firstname.lastname@univ-lille.fr
}
}

\maketitle

\begin{abstract}
In recent years, spiking neural networks (SNNs) emerge as an alternative to deep  neural  networks  (DNNs). SNNs present a higher computational efficiency -- using low-power neuromorphic hardware -- and require less labeled data for training -- using local and unsupervised learning rules such as spike timing-
dependent plasticity (STDP). SNN have proven their effectiveness in image classification on simple datasets such as MNIST. However, to process natural images, a pre-processing step is required. Difference-of-Gaussians (DoG) filtering is typically used together with on-center~/~off-center coding, but it results in a loss of information that is detrimental to the classification performance. In this paper, we propose to use whitening as a pre-processing step before learning features with STDP. Experiments on CIFAR-10 show that whitening allows STDP to learn visual features that are closer to the ones learned with standard neural networks, with a significantly increased classification performance as compared to DoG filtering. We also propose an approximation of whitening as convolution kernels that is computationally cheaper to learn and more suited to be implemented on neuromorphic hardware. Experiments on CIFAR-10 show that it performs similarly to regular whitening. Cross-dataset experiments on CIFAR-10 and STL-10 also show that it is fairly stable across datasets, making it possible to learn a single whitening transformation to process different datasets.
\end{abstract}


\section{Introduction}

In recent years, \glspl{dnn} have become a \emph{de facto} standard in machine learning, thanks to their ability to learn complex representations from large amounts of data. They have demonstrated their superiority over other models in a large number of tasks, including image and video classification, speech recognition, and natural language understanding. However, they suffer from two major drawbacks that hamper their adoption at a large scale. First, training a \gls{dnn} is computationally expensive, due to its large number of parameters and the large amounts of data required to effectively estimate these parameters. As a consequence, \glspl{dnn} training is usually performed on GPU or TPU that consume large quantities of energy. Moreover, \glspl{dnn} mostly rely on supervised learning, which requires these large amounts of data ({e.g.}, millions of samples in the case of image classification~\cite{krizhevsky12imagenet}) to be annotated manually beforehand, making them difficult to apply to new tasks, unless one is willing to spend large amounts of time and money on the labeling process. \Glspl{snn} offer an alternative to \glspl{dnn}; they can be implemented efficiently through low-power neuromorphic hardware~\cite{merolla2014million}, which would solve one issue of \glspl{dnn}. The problem of data labeling can also be avoided -- to some extent -- through the use of unsupervised learning rules. \Gls{stdp} is one of those rules, that can enable effective unsupervised learning in \glspl{snn}~\cite{masquelier08spike} and is compatible with low-energy neuromorphic hardware~\cite{schuman2017survey}.

In this paper, we are interested in the problem of visual feature learning through \glspl{snn} equipped with an \gls{stdp} learning rule, with the long-term goal of producing end-to-end spiking architectures compatible with low-power, neuromorphic, hardware. Such a system includes the following steps: image pre-processing, neural coding of the pre-processed images into spikes, neuron and synapse models, learning rules, and finally the feature classifier; all these elements should ideally be implementable through neuromorphic hardware components.  In~\cite{falez2019unsupervised}, an in-depth study of \gls{stdp}-based feature learning for image recognition was performed; it concluded that \gls{stdp}-based \glspl{snn} cannot currently compete with more traditional neural networks models of visual feature learning (namely, auto-encoders), and pointed out some reasons for the ineffectiveness of \glspl{snn}, especially the pre-processing of images and the inhibition mechanisms.

\begin{figure*}[ht]
\centering
\subfloat[]{\includegraphics[width=0.15\textwidth]{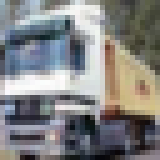}}
\qquad
\subfloat[]{\includegraphics[width=0.15\textwidth]{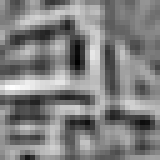}}
\qquad
\subfloat[]{\includegraphics[width=0.15\textwidth]{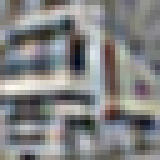}}
\qquad
\subfloat[]{\includegraphics[width=0.15\textwidth]{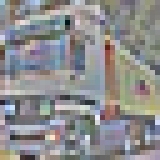}}
\caption{Pre-processing for color images: (a) raw RGB image, (b) on-center~/~off-center coding on grayscale image (c) on-center~/~off-center coding on color image, and (d) whitened RGB image.}
\label{figure:intro:pre-processing}
\end{figure*}

In this paper, we specifically address the issue of image pre-processing for visual feature learning and natural image classification with \gls{stdp}-based \glspl{snn}. \gls{stdp} learns patterns of correlated spike timestamps from the input spike trains~\cite{masquelier08spike}. To be processed by \gls{stdp} networks, natural images must first be pre-processed so that the spikes trains representing them can encode relevant visual information. Indeed, directly encoding pixel values as spikes, through either temporal or frequency coding, would lead to learning mostly patterns consisting of uniform regions of light colors; this has been empirically confirmed in~\cite{falez2019unsupervised} in the case of temporal coding. The most common way to circumvent this issue is to convert images to grayscale and to apply on-center~/~off-center coding~\cite{delorme2001networks,kheradpisheh2018stdp} (or some equivalent edge-extraction method such as Gabor filters~\cite{kheradpisheh2016bio}). This coding is inspired by biological vision, and is also related to the SIFT keypoint detector widely used in computer vision~\cite{lowe2004distinctive}. It extracts edges from the images by applying a \gls{dog} filter (see Figure~\ref{figure:intro:pre-processing}(b)). As a result, the spike trains now encode edge information, which is richer than raw pixel information. However, this prevents the \gls{stdp} networks from learning also visual patterns based on colors, as standard deep neural networks do~\cite{coates2011analysis,krizhevsky12imagenet}. Applying on-center~/~off-center coding to the R, G, and B color channels independently (see Figure~\ref{figure:intro:pre-processing}) does not solve this issue, since it only allows to learn edge patterns specific to one of the three color channels rather than actual color patterns. It has been shown empirically in~\cite{falez2019unsupervised} that on-center~/~off-center coding, applied either to grayscale or color images, results in a loss of information that is detrimental to image classification.

In this work, we propose to use whitening as a single pre-processing step for processing natural images with \gls{stdp}-based \glspl{snn}. Whitening is commonly used in computer vision as an image pre-processing method~\cite{krizhevsky2009learning,coates2011analysis}, among other uses~\cite{huang18decorrelated}. Generally speaking, whitening is a procedure used in statistics to standardize and de-correlate data; it projects the data into a new, orthonormal, space so that its components are centered, independent, and have unit variance. When applied to images, whitening discards first-order correlations, which correspond to the fact that pixels that are spatially close to each other tend to have similar values; visually, it highlights edges and high frequency features. It allows neural networks to learn non-trivial correlations between pixels~\cite{coates2011analysis}.

The whitening transformation is typically computed from a dataset using principal component analysis (\gls{pca}) or \gls{zca}~\cite{bell97a}, and applied to whole images. As we aim at training end-to-end \gls{stdp}-based systems that can be implemented on energy-efficient hardware, whitening cannot be applied as is, for three reasons:
\begin{itemize}
    \item learning a \gls{pca} or \gls{zca} transformation on whole images is computationally expensive and involves operations on dense matrices that cannot be implemented simply through neuromorphic hardware;
    \item only applying a learned whitening transformation to whole images cannot be implemented efficiently through neuromorphic hardware either, because of the non-local nature of the transformation;
    \item the transformation is data-dependent, so a new transformation should be computed for every new dataset.
\end{itemize}
The use of whitening as a pre-processing step for \gls{snn}-based image analysis would only be valuable if these issues can be circumvented in some way.

Based on these observations, the contribution of this paper is three-fold.
\begin{enumerate}
  \item We show that using whitening as a pre-processing step allows \gls{stdp} to learn patterns that are similar to what standard deep neural networks can learn, and is superior to on-center~/~off-center coding when performing image classification based on features learned with \gls{stdp}.
  \item We propose an approximation of \gls{zca} whitening based on convolution kernels, that can be pre-computed more efficiently and fits the constraints of neuromorphic hardware better. Experiments show that this approximation yields the same performances as standard \gls{zca} whitening.
  \item We show through cross-dataset experiments that it is possible to pre-compute a single whitening transformation on one dataset and apply it to other datasets with only little impact on the classification performance. The resulting transformation still outperforms significantly on-center~/~off-center coding.
\end{enumerate}




\section{Related Work}

\paragraph{SNN-based Visual Feature Learning}
Image classification and visual feature learning with SNNs has received an increasing interest over the last years (see~\cite{tavanaei19a} for a recent survey on this topic). Most authors focus on simple datasets like MNIST, which offer limited challenges. Some models were evaluated on more complex datasets of natural images such as CIFAR-10, but they usually include training procedures that cannot be implemented on low-power neuromorphic hardware, the most common approach being to convert pre-trained DNN models to spiking models. Such models offer limited benefits over DNNs, since training the model is the most computationally expensive step. SNN models that can be trained on energy-efficient hardware are usually based on STDP learning rules. The performance of these models are still behind other models, especially DNNs; as a consequence, most work still focus on simple datasets, such as MNIST~\cite{querlioz2011simulation, diehl2015unsupervised, tavanaei2016bio} and ETH-80~\cite{kheradpisheh2016bio, kheradpisheh2018stdp}. This may be due to the difficulty of training multi-layer STDP networks: multi-layer models based on STDP~\cite{kheradpisheh2018stdp,falez2019unsupervised} have only be proposed very recently. Another reason is the difficulty to handle color natural images, due to the ineffectiveness of the on-center~/~off-center coding used to pre-process images~\cite{falez2019unsupervised}. As a result, complex datasets of natural images are seldom used to evaluate STDP-based SNNs; recent examples include the Caltech Faces/Motorbikes dataset~\cite{kheradpisheh2018stdp,falez2019multi}, and CIFAR-10, CIFAR-100 or STL-10~\cite{falez2019unsupervised}. In this paper, we aim at improving the ability of STDP-based networks to learn from complex natural images and bringing their performance closer to their competitors.

\paragraph{Whitening for Feature Learning and Deep Learning}
Whitening has been especially studied as a pre-processing step in unsupervised visual feature learning~\cite{krizhevsky2009learning,coates2011analysis}. The reported results were sometimes contradictory: Krizhevsky and Hinton~\cite{krizhevsky2009learning} evaluated the role of whitening in image classification based either on raw pixels or on visual features learned with restricted Boltzmann machines (RBM), and concluded whitened images did not provide any improvement over using unwhitened images, whereas Coates~\emph{et al.}~\cite{coates2011analysis} reported significant and consistent improvements in classification performance when using whitened images to learn visual features with k-means, mixtures of Gaussians, auto-encoders, and RBM. Whatever the actual outcomes of whitening can be when using traditional algorithms, the debate is not relevant with STDP-based SNNs, as learning from raw images is not an option in this case, as demonstrated in~\cite{falez2019unsupervised}. In addition to pre-processing input data, whitening can be used to normalize network activations between layers of a \gls{dnn}~\cite{huang18decorrelated}, in a process similar to batch normalization. However, this is not related to our goal here, which is only to consider whitening as an alternative to on-center~/~off-center coding.

\paragraph{Whitening and SNNs} 
Whitening is seldom used as a pre-processing step to \gls{snn}. To our knowledge, only Burbank~\cite{burbank2015mirrored} used whitening as a pre-processing step before learning visual features from natural images with \glspl{snn}. No specific reason for the use of whitening was provided, other than re-using the data of Olshausen \& Field~\cite{olshausen96emergence}. The evaluation of the resulting features does not include recognition performance and the performance of whitening w.r.t. other pre-processing methods (e.g., on-center~/~off-center coding) was not assessed.

\section{Background}

\subsection{Unsupervised Feature Learning and Image Classification}
The problem of image classification can be modeled as finding a function~\(f: \mathbb{I} \rightarrow \mathbb{C}\) which assigns to an image \(I \in \mathbb{I}\) the label \(c \in \mathbb{C}\) of the class it belongs to. A typical DNN-based approach would directly infer \(f\) from labeled training data. Other approaches  model \(f\) as the composition of three individual functions: a feature extractor \(f_e\), a feature aggregator \(f_a\), and a supervised classifier \(f_c\). The feature extractor is a function \(f_e: \mathbb{I} \rightarrow \mathbb{R}^{m \times d}\) that converts an image \(I\) into a set of \(m\) visual features representative of its visual content (shape, color\ldots); each feature is modeled as a vector of dimension \(d\). 
The feature aggregator is a function \(f_a: \mathbb{R}^{m \times d} \rightarrow \mathbb{R}^{d'}\) that aggregates the \(m\) feature vectors into a single description vector of dimension \(d'\), typically through some pooling operation. 
Finally, the classifier \(f_c: \mathbb{R}^{d'} \rightarrow \mathbb{C}\) assigns a class \(c \in \mathbb{C}\) to an image \(I \in \mathbb{I}
\) based on its aggregated feature vector \(f_a \circ f_e(I))\)~: \(c = f_c \circ f_a \circ f_e(I)\). 

In this work, the feature extractor \(f_e\) is a convolutional single-layer SNN that learns features from data using an unsupervised STDP learning rule. As we aim at evaluating only the ability of STDP to learn visual features, we rely on more classical tools for the feature aggregator \(f_a\) (max pooling) and the classifier \(f_c\) (SVM). The details of our recognition system are provided in Section~\ref{sec:results:recognition}.

\subsection{SNN model}
\label{section:snn_model}
An \gls{stdp}-based \gls{snn} typically includes the following components: a neural coding model, which converts input data into spikes, a spiking neuron model, an \gls{stdp} learning rule, and homeostasis mechanisms that ensure that the activity of the network remains consistent. Since we are interested in learning visual features that will be use for classification, we also need a "neural decoding" model that converts output spikes back to numerical values that can be fed to the feature aggregator or the classifier. We use the same components as in  of~\cite{falez2019multi}, which provide state-of-the-art performance for \gls{stdp}-based visual feature learning.

\paragraph{Neural coding}
We use latency coding~\cite{thorpe2001spike}, which is one variant of temporal coding, to convert input values \gls{m_input_value} into spikes. Earlier spikes encode larger values. Spike timestamps are generated as follows:
\begin{equation}
\gls{m_time} = \gls{m_exposition_duration}(1-\gls{m_input_value})
\label{eq:latency_coding}
\end{equation}
with \gls{m_time} the timestamp of the spike, \gls{m_input_value} the converted input value, and \gls{m_exposition_duration} the duration of the exposition of an input sample to the network. As a consequence, there is at most one spike per input per sample.

\paragraph{Neuron model} The \gls{snn} uses integrate-and-fire (\gls{if}) neurons~\cite{burkitt2006review}. This model is defined as follows:
\begin{equation}
 \gls{m_membrane_capacitance}\frac{\partial\gls{m_neuron_potential}}{\partial\gls{m_time}} = \gls{m_input_current}(\gls{m_time}), \gls{m_neuron_potential}\leftarrow \gls{m_resting_potential}~\text{when}~\gls{m_neuron_potential}\geq\gls{m_potential_threshold}
\label{eq:if}
\end{equation}
with \gls{m_neuron_potential} the membrane potential, \gls{m_resting_potential} the resting potential, \gls{m_membrane_capacitance} the membrane capacitance, \gls{m_potential_threshold} the threshold of the neuron, and \(\gls{m_input_current}(\gls{m_time})\) the input current of the neuron (\(\gls{m_input_current}(\gls{m_time}) = 1\) if an input spike is received at time \(t\), and \(\gls{m_input_current}(\gls{m_time}) = 0\) otherwise).

\paragraph{Synapse model}
Every time a neuron fires a spike, the weights of its input connections are updated following a STDP rule, according to the activity of the corresponding pre-synaptic neurons. Multiplicative \gls{stdp}~\cite{querlioz2011simulation} is used to train synaptic weights \gls{m_weight}:
\begin{equation}
  \gls{m_weight_change} = \left \{
   \begin{array}{r l}
     \gls{m_weight_lr}e^{-\gls{m_stdp_beta}\frac{\gls{m_weight}-\gls{m_min_weight}}{\gls{m_max_weight}-\gls{m_min_weight}}} & \text{if}~\gls{m_time_pre}\leq\gls{m_time_post}\\
     & \text{and}~\gls{m_time_post}-\gls{m_time_pre}\leq\gls{m_ltp_window} \\
     & \\
     -\gls{m_weight_lr}e^{-\gls{m_stdp_beta}\frac{\gls{m_max_weight}-\gls{m_weight}}{\gls{m_max_weight}-\gls{m_min_weight}}} & \text{otherwise}
   \end{array}
   \right .
\label{eq:multiplicative_stdp}
\end{equation}
with \gls{m_min_weight} and \gls{m_max_weight} the bounds of the weight \gls{m_weight}, \gls{m_weight_change} the update applied the weight (\(w_{t_1} = w_t + \Delta_w\)), \gls{m_weight_lr} the learning rate, and \gls{m_time_pre} and \gls{m_time_post} the firing timestamps of the pre-synaptic and post-synaptic neurons, respectively. \gls{m_stdp_beta} is a parameter that controls the saturation effect of the learning rule (increasing~\gls{m_stdp_beta} reduces the saturation of weights).

\paragraph{Homeostasis}
Homeostasis in the network is maintained through the adaptation of neuron thresholds. Threshold values \gls{m_potential_threshold} are learned with the threshold adaptation proposed in~\cite{falez2019multi} and a \gls{wta} mechanism. Under this model, when a neuron wins the competition (i.e. it is the first one to fire a spike during the exposition of a sample), it applies the \gls{stdp} rule to update its synaptic weights and it adapts its threshold \gls{m_potential_threshold} with the following update:
\begin{equation}
    \glslink{m_threshold_change}{\Delta_{\operatorname{th}}^1} = \gls{m_threshold_lr},
\label{eq:th_wta_win}
\end{equation}
with $\Delta_{\operatorname{th}}^1$ the change applied to the neuron threshold \gls{m_potential_threshold} and \gls{m_threshold_lr} the learning rate of threshold adaptation. This rule ensures that a neuron will not always be the first to emit a spike.

The others neurons do not apply \gls{stdp} and decrease their threshold as follows:
\begin{equation}
    \glslink{m_threshold_change}{\Delta_{\operatorname{th}}^1} = -\frac{\gls{m_threshold_lr}}{|\gls{m_layer_output}|}
\label{eq:th_wta_lose}
\end{equation}
with $|\gls{m_layer_output}|$ the number of neurons in competition. This second update promotes diversity in neurons by lowering the threshold of neurons that emit few or no spikes.

Moreover, each time a winning neuron fires a spike, all the neurons in competition apply the following update to their threshold:
\begin{equation}
     \glslink{m_threshold_change}{\Delta_{\operatorname{th}}^2} = -\gls{m_threshold_lr}(\gls{m_time}-\gls{m_expected_time})
     \label{eq:th_target}
\end{equation}
with \gls{m_threshold_lr} the threshold learning rate and \gls{m_time} the timestamp at which the neuron fired the spike. \Gls{m_expected_time} is a manually-defined timestamp objective at which neurons should fire spikes. This parameter controls the number of input spikes to be integrated before an output spike is emitted, and, so, the nature of the filters to be learned~\cite{falez2019unsupervised,falez2019multi}.

\paragraph{Neural decoding}
Spikes generated at the output of the \gls{snn} can be converted back into values as follows:
\begin{equation}
    \gls{m_output_value} = \min\left(1, \max\left(0, 1-\frac{t-\gls{m_expected_time}}{\gls{m_exposition_duration}-\gls{m_expected_time}}\right)\right)
    \label{eq:coding_inv}
\end{equation}

\subsection{ZCA Whitening}
\label{ssec:zca_whitening}
Whitening is a data-dependent transformation that decorrelates and standardizes the data. Several whitening transformations can exist for a given dataset, as whitened data remains whitened under rotations. Among these, zero-phase whitening (ZCA)~\cite{bell97a} is the transformation that produces the whitened data that remains the closest to the original data. When applied to images, ZCA whitening produces images that are still recognizable by the human eye (as opposed to, for instance, PCA whitening).

Let \(X\) be a centered data matrix and \gls{m_cov_matrix} its covariance matrix. \gls{m_cov_matrix} can be decomposed so that:
\begin{equation}
  \gls{m_cov_matrix} = \gls{m_eigen_vector_matrix}\gls{m_eigen_value_matrix}\gls{m_eigen_vector_matrix}^{-1}
\end{equation}
with~\gls{m_eigen_vector_matrix} the matrix of eigenvectors of \gls{m_cov_matrix} and~\gls{m_eigen_value_matrix} the diagonal matrix of its eigenvalues ($\gls{m_eigen_value_matrix} = \text{diag}(\gls{m_eigen_value}_1, \gls{m_eigen_value}_2, \ldots, \gls{m_eigen_value}_n)$).

The ZCA transformation matrix \gls{m_whiten_matrix} for data \(X\)  is computed as follows:
\begin{equation}
  \gls{m_whiten_matrix} =  \gls{m_eigen_vector_matrix}\sqrt{(\gls{m_eigen_value_matrix}+\gls{m_whiten_coef})^{-1}}\gls{m_eigen_vector_matrix}^T
  \label{eq:whitening}
\end{equation}
with~\gls{m_whiten_coef} the whitening coefficient, which adds numerical stability and acts as a low pass filter. As in PCA, it is possible to retain only the \(k\) largest eigenvalues and their corresponding eigenvectors, to eliminate the least significant components of the data, which may correspond to noise. In the remaining of the paper, we note $\gls{m_eigen_ratio}\in[0, 1]$ the ratio of the largest eigenvalues that are retained.

Finally, the ZCA whitened data \gls{m_image_whiten} is computed from the ZCA transformation \gls{m_whiten_matrix} as:
\begin{equation}
  \gls{m_image_whiten} = \gls{m_whiten_matrix}\gls{m_image}
\end{equation}

\section{Contribution}

\subsection{Encoding Whitened Data as Spikes}
\label{ssec:trad_whitening}

The first part of our contribution is to enable the conversion of whitened data into spikes. The output of the whitening transformation contain both positive and negative values, which correspond to the positive or negative contribution of the data to the components of the transformation; larger values correspond to larger contributions, i.e. they are more significant. These values must be converted into spike while respecting the principle of temporal coding: larger values must be converted into the earlier spikes. Similarly to on-center~/~off-center coding, we split the values into two channels, a positive one and a negative one. The conversion process follows these steps:
\begin{enumerate}
    \item The data matrix \gls{m_image} is whitened using the learned \gls{zca} transformation.
    \item The components of each sample in \gls{m_image_whiten} are scaled in $[-1, 1]$ according to the minimum and maximum values of the sample.
    \item Positive and negative values are split into two channels \gls{m_pos_value} and \gls{m_neg_value}: $\gls{m_pos_value} = \max(0, \gls{m_image_whiten})$, $\gls{m_neg_value} = \max(0, -\gls{m_image_whiten})$.
\end{enumerate}
The values can finally be converted into spikes by using latency coding (Equation~\ref{eq:latency_coding}).


\subsection{Approximating whitening with convolution kernels}
\label{ssec:patch_whitening}
Applying the whitening transformation to images is computationally expensive and is not easily implementable on neuromorphic architectures. In opposition, the DoG filter of on-center~/~off-center coding is a pre-processing, which is already well-used with \glspl{snn}, can be computed by simply convolving an image with an appropriate kernel. 
In this section, we show how to approximate whitening by convolution kernels, to benefit both from the ease of implementation of DoG filtering and from the performance of whitening. Our approach also reduces the cost of learning the whitening transformation matrix.

The general idea is to learn the whitening transform on small patches rather whole images, then to approximate the patch whitening transformation by the whitening transformation of a single pixel within the patches, which can be expressed as a convolution kernel. First, \gls{m_patch_number} patches of size  $\gls{m_patch_width}\times\gls{m_patch_height}$ are extracted from the dataset (e.g. using dense or random sampling). A \gls{zca} transformation matrix \gls{m_whiten_matrix} of dimension $[\gls{m_patch_width}\times\gls{m_patch_height}\times\gls{m_image_depth}, \gls{m_patch_width}\times\gls{m_patch_height}\times\gls{m_image_depth}]$ is computed from these patches. Finally, \gls{m_whiten_matrix} is converted into \gls{m_image_depth} kernels \(K_c\) of dimension $[\gls{m_patch_width}, \gls{m_patch_height}]$. To do so, an impulse response matrix \gls{m_pulse_martix} is created for each \gls{m_image_depth} channel with only its central value in channel \(c\) set to $1$:
\begin{equation}
    \gls{m_pulse_martix}_c(i,j,k) = \left \{
     \begin{array}{r l}
      1 & \text{if}~k=c~\text{and}~i=\frac{\gls{m_patch_width}}{2}~\text{and}~j=\frac{\gls{m_patch_height}}{2}  \\
      0 & \text{otherwise}
     \end{array}
     \right.
\end{equation}
with $i\in[0, \gls{m_patch_width}]$, $j\in[0, \gls{m_patch_height}]$, and $k\in[0, \gls{m_image_depth}]$ the coordinates in matrix \gls{m_pulse_martix} and $c\in[0, \gls{m_image_depth}]$ the corresponding channel.

\begin{figure}
  \centering
  \subfloat[$\gls{m_patch_width},\gls{m_patch_height}=5\times5$, $\gls{m_eigen_ratio} = 0.2$, $\gls{m_whiten_coef} = 10^{-3}$.]{
      \begin{tabular}{cccc}
        \includegraphics[width=0.18\columnwidth]{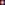} &
        \includegraphics[width=0.18\columnwidth]{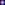} &
        \includegraphics[width=0.18\columnwidth]{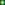} &
        \includegraphics[width=0.18\columnwidth]{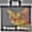} \\
      \end{tabular}
    }

    \subfloat[$\gls{m_patch_width},\gls{m_patch_height}=9\times9$, $\gls{m_eigen_ratio} = 0.6$, $\gls{m_whiten_coef} = 10^{-3}$.]{
        \begin{tabular}{cccc}
          \includegraphics[width=0.18\columnwidth]{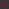} &
          \includegraphics[width=0.18\columnwidth]{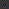} &
          \includegraphics[width=0.18\columnwidth]{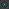} &
          \includegraphics[width=0.18\columnwidth]{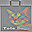} \\
        \end{tabular}
      }

      \subfloat[$\gls{m_patch_width},\gls{m_patch_height}=11\times11$, $\gls{m_eigen_ratio} = 1.0$, $\gls{m_whiten_coef} = 10^{-3}$.]{
          \begin{tabular}{cccc}
            \includegraphics[width=0.18\columnwidth]{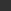} &
            \includegraphics[width=0.18\columnwidth]{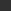} &
            \includegraphics[width=0.18\columnwidth]{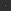} &
            \includegraphics[width=0.18\columnwidth]{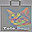} \\
          \end{tabular}
        }

        \subfloat[$\gls{m_patch_width},\gls{m_patch_height}=9\times9$, $\gls{m_eigen_ratio} = 1.0$, $\gls{m_whiten_coef} = 10^{-1}$.]{
            \begin{tabular}{cccc}
              \includegraphics[width=0.18\columnwidth]{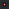} &
              \includegraphics[width=0.18\columnwidth]{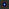} &
              \includegraphics[width=0.18\columnwidth]{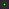} &
              \includegraphics[width=0.18\columnwidth]{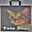} \\
            \end{tabular}
          }
  \caption{Examples of whitening kernels approximating the \gls{zca} transformation.}
  \label{fig:zca_filter}
\end{figure}

A whitening kernel can be computed for each channel \(c\) as:
\begin{equation}
    K_c = \gls{m_pulse_martix}_c \gls{m_whiten_matrix}
\end{equation}
and an image can be whitened by convolving each of its channels with the corresponding whitening kernel. Each whitening kernel \(K_c\) corresponds to the whitening transformation of the central pixel of a channel of the patches. Examples of whitening kernels generated by this method and the resulting filtered images are shown in Figure~\ref{fig:zca_filter}.

\section{Results}

\subsection{Experimental Protocol}
\label{sec:results:recognition}

\paragraph{Objectives}
In these experiments, we evaluate:
\begin{itemize}
    \item the performance of our whitening kernels versus standard whitening, and its sensitivity to major parameters;
    \item the performance of whitening as a pre-processing step for \gls{stdp}-based visual feature learning versus standard on-center~/~off-center coding;
    \item the stability of whitening kernels across datasets, by performing cross-dataset experiments in which the whitening transformation is trained on one dataset and applied to another dataset to perform feature learning and image recognition.
\end{itemize}

\paragraph{Recognition system}

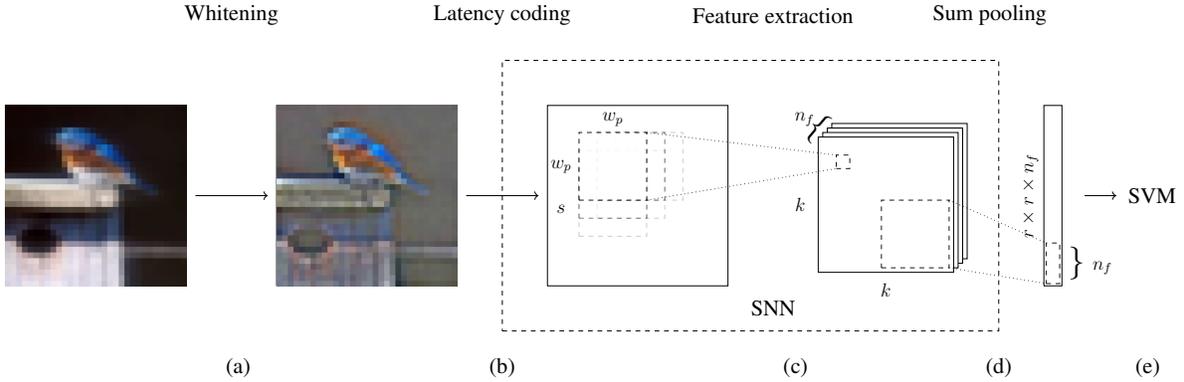
\begin{figure*}
    \centering
    \scalebox{0.6}{
        \usetikzlibrary{calc}

\newcommand{\rectangle}[4]
{ 
	\draw[fill=white] (#1,#2) -- (#1+#3,#2) -- (#1+#3,#2+#4) -- (#1,#2+#4) --  (#1,#2) ;
}

\newcommand{\dashedrectangle}[5]
{ 
	\draw[fill=white, dashed,opacity=#5] (#1,#2) -- (#1+#3,#2) -- (#1+#3,#2+#4) -- (#1,#2+#4) --  (#1,#2) ;
}

\begin{tikzpicture}

	\def\imageax{-12}
	\def\imageay{0}

	\def\imagebx{-6}
	\def\imageby{0}

	\def\imagex{0}
	\def\imagey{0}
	\def\imagew{4}
	\def\imageh{4}

	\def\patchi{1.5}
	\def\patchix{0.7}
	\def\patchiy{0.6}
	\def\patchis{0.4}

	\def\convx{6}
	\def\convy{0.3}
	\def\convw{3}
	\def\convh{3}

	\def\patchc{0.3}
	\def\patchcx{0.4}
	\def\patchcy{0.4}

   \def\patchd{1.5}
	\def\patchdx{1.4}
	\def\patchdy{1.4}

	\def\poolx{11}
	\def\pooly{0}
	\def\poolw{0.4}
	\def\poolh{4}

   \def\patchpw{0.3}
   \def\patchph{0.9}
	\def\patchpx{0.05}
	\def\patchpy{3.05}

	\def\depthf{0.1}

	\rectangle{\imageax+0*\depthf}{\imageay+0*\depthf}{\imagew}{\imageh}
	
	\path[fill overzoom image=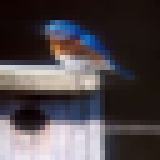] (\imageax, \imageay) rectangle (\imageax+\imagew, \imageay+\imageh);

	\rectangle{\imagebx+0*\depthf}{\imageby+0*\depthf}{\imagew}{\imageh}
	
	\path[fill overzoom image=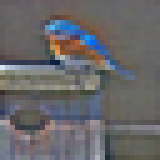] (\imagebx, \imageby) rectangle (\imagebx+\imagew, \imageby+\imageh);
	\draw[->] (\imageax+\imagew+0.2, \imageby+\imageh/2) -- (\imagebx-0.2, \imageby+\imageh/2);
	\dashedrectangle{\imagex-1}{\imagey-1}{11}{6}{1.0}

	\node at (-7, 6) {\Large Whitening};
	\draw[->] (\imagebx+\imagew+0.2, \imagey+\imageh/2) -- (\imagex-0.2, \imagey+\imageh/2);
	\rectangle{\imagex+0*\depthf}{\imagey+0*\depthf}{\imagew}{\imageh}

	\node at (-1, 6) {\Large Latency coding};

	\node at (-6.85, -1.8) {\Large (a)};

	
	\dashedrectangle{\imagex+\patchix+1*\patchis}{\imagey+\imageh-\patchi-\patchiy-1*\patchis}{\patchi}{\patchi}{0.2}
	\dashedrectangle{\imagex+\patchix+0*\patchis}{\imagey+\imageh-\patchi-\patchiy-2*\patchis}{\patchi}{\patchi}{0.2}
	\dashedrectangle{\imagex+\patchix+2*\patchis}{\imagey+\imageh-\patchi-\patchiy-0*\patchis}{\patchi}{\patchi}{0.2}
	\dashedrectangle{\imagex+\patchix+0*\patchis}{\imagey+\imageh-\patchi-\patchiy-1*\patchis}{\patchi}{\patchi}{0.5}
	\dashedrectangle{\imagex+\patchix+1*\patchis}{\imagey+\imageh-\patchi-\patchiy-0*\patchis}{\patchi}{\patchi}{0.5}
	\dashedrectangle{\imagex+\patchix+0*\patchis}{\imagey+\imageh-\patchi-\patchiy-0*\patchis}{\patchi}{\patchi}{0.9}


	\node at (\imagex+\patchix-0.4, \imagey+\imageh-\patchiy-\patchi/2) {\large $w_p$};
	\node at (\imagex+\patchix+\patchi/2, \imagey+\imageh-\patchiy+0.3) {\large $w_p$};

	\node at (\imagex+\patchix-0.4, \imagey+\imageh-\patchiy-\patchi-\patchis/2) {\large $s$};

	\node at (5, 6) {\Large Feature extraction};

	\node at (-1, -1.8) {\Large (b)};

	\rectangle{\convx+3*\depthf}{\convy+3*\depthf}{\convw}{\convh}
	\rectangle{\convx+2*\depthf}{\convy+2*\depthf}{\convw}{\convh}
	\rectangle{\convx+1*\depthf}{\convy+1*\depthf}{\convw}{\convh}
	\rectangle{\convx+0*\depthf}{\convy+0*\depthf}{\convw}{\convh}

	\dashedrectangle{\convx+\patchcx}{\convy+\convh-\patchc-\patchcy}{\patchc}{\patchc}{0.9}
	\draw[dotted] (\imagex+\patchi+\patchix, \imagey+\imageh-\patchiy) -- (\convx+\patchcx, \convy+\convh-\patchcy);
	\draw[dotted] (\imagex+\patchi+\patchix, \imagey+\imageh-\patchiy-\patchi) -- (\convx+\patchcx, \convy+\convh-\patchcy-\patchc);

	\node at (\convx-0.4, \convy+\convh/2) {\large $k$};
	\node at (\convx+\convw/2, \convy-0.4) {\large $k$};

	\node [rotate=50+90+180] at (\convx, \convy+\convh+0.2) {\huge $\left\{ \right.$};
	\node at (\convx-0.3, \convy+\convh+0.4) {\large $n_f$};

	\node at (9.8, 6) {\Large Sum pooling};
	\node at (5.5, -1.8) {\Large (c)};
	\rectangle{\poolx+0*\depthf}{\pooly+0*\depthf}{\poolw}{\poolh}

	\dashedrectangle{\convx+\patchdx}{\convy+\convh-\patchd-\patchdy}{\patchd}{\patchd}{0.9}
	\dashedrectangle{\poolx+\patchpx}{\pooly+\poolh-\patchph-\patchpy}{\patchpw}{\patchph}{0.9}
	\draw[dotted] (\convx+\patchdx+\patchd,\convy+\convh-\patchd-\patchdy) -- (\poolx+\patchpx, \pooly+\poolh-\patchph-\patchpy);
	\draw[dotted] (\convx+\patchdx+\patchd,\convy+\convh-\patchd-\patchdy+\patchd) -- (\poolx+\patchpx, \pooly+\poolh-\patchph-\patchpy+\patchph);

	\node[rotate=90] at (\poolx-0.3, \pooly+\poolh/2) {\large $r \times r \times n_f$};
    
  \node [rotate=180] at (\poolx+\poolw+0.3, \pooly+\poolh-\patchpy-\patchph/2) {\huge $\left\{ \right.$};
  
    \node at (\poolx+\poolw+0.9, \pooly+\poolh-\patchpy-\patchph/2-0.1) {\large $n_f$};

	\draw[->] (\poolx+\poolw+0.5,\pooly+\poolh/2) -- (\poolx+\poolw+1.2,\pooly+\poolh/2);
	\node at (\poolx+\poolw+2,\pooly+\poolh/2) {\Large SVM};
	\node at (10, -1.8) {\Large (d)};
	\node at (13.3, -1.8) {\Large (e)};
	\node at (5.0, -0.5) {\Large SNN};
	
\end{tikzpicture}
    }
    \caption{Recognition system used in our experiments.}
    \label{fig:protocol}
\end{figure*}

Our recognition system follows the same general procedure as \cite{coates2011analysis} and \cite{falez2019unsupervised}. The system is depicted in Figure~\ref{fig:protocol}. The major stages of the system are:
\begin{enumerate}
    \item Image pre-processing (Figure~\ref{fig:protocol}(a)) through on-center~/~off-center coding, standard whitening (see Section~\ref{ssec:zca_whitening}), or whitening kernels (see Section~\ref{ssec:patch_whitening}).
    \item Feature extraction (\(f_e\)) by a single-layer convolutional \gls{snn} following the model presented in Section~\ref{section:snn_model} (see Figure~\ref{fig:protocol}(c)).
    \item Feature aggregation (\(f_a\)) through sum pooling over $2\times2$ non-overlapping image regions (see Figure~\ref{fig:protocol}(d)).
    \item Feature vector classification (\(f_c\)) with a linear SVM (see Figure~\ref{fig:protocol}(e)).
\end{enumerate}

\paragraph{Datasets}
We use CIFAR-10~\cite{krizhevsky2009learning} as a reference dataset in our experiments. This dataset 60,000 color images of size $32\times32$, divided into 10 classes; it is split into a training set of 50,000 images and a test set of 10,000 images. We also use the labeled part of STL-10~\cite{coates2011analysis} for cross-dataset experiments. It contains 13,000 \(96\times 96\) color images split into 5,000 training images and 8,000 test images. Note that the scales of the images in the two datasets are different, making cross-dataset experiments more challenging for our whitening kernels.

\paragraph{Computation of the whitening transformations}
Whitening transformations are computed on the training set of CIFAR-10 for regular experiments; for cross-dataset experiments, they are computed on STL-10 (resp. CIFAR-10) when feature learning and classification is performed on CIFAR-10 (resp. STL-10). The standard whitening transformation is learned using the whole training set. Patch-based whitening transformations are learned on $10^6$ patches densely sampled with a stride of 2 from images of the training set.

\paragraph{Implementation Details}
The parameters of the \gls{snn} are set to the values in Table~\ref{table:parameters}, unless otherwise specified. All configurations are run 3 times; average recognition rates over the three runs and their standard deviations are reported. \gls{csnns}\footnote{This tool is open-source and can be downloaded at \url{https://gitlab.univ-lille.fr/bioinsp/falez-csnn-simulator}}~\cite{falez2019improving} is used to implement all the experiments.

\begin{table}
\centering
\begin{tabular}{lrlr}
    \toprule
    \multicolumn{4}{c}{\textbf{Neural Coding}}\\
    \gls{m_exposition_duration} & $1$ & &\\
    \midrule
    \multicolumn{4}{c}{\textbf{Neuron}}\\
    $\gls{m_potential_threshold}(0)$ & $\sim\gls{m_normal_distribution}(10, 0.1)$ & \gls{m_resting_potential} & $0$\\
    \midrule
    \multicolumn{4}{c}{\textbf{Threshold Adaptation}}\\
    \gls{m_expected_time} & $0.97$ & $\gls{m_threshold_lr}(0)$ & $1$\\
    \midrule
    \multicolumn{4}{c}{\textbf{Training}}\\
    \gls{m_annealing} & $0.95$ & \gls{m_epoch_number} & $100$\\
    \midrule
    \multicolumn{4}{c}{\textbf{STDP}}\\
    \gls{m_min_weight} & $0$ & \gls{m_max_weight} & $1$ \\
    $\gls{m_weight_lr}(0)$ & $0.1$ & $\gls{m_weight}(0)$ & $\sim\gls{m_uniform_distribution}(0, 1)$\\
    \gls{m_stdp_beta} & $1$ & & \\
    \midrule
    \multicolumn{4}{c}{\textbf{Network architecture}}\\
    filter size & $5\times5$ & stride & $1$ \\
    padding & $0$ & & \\
    \bottomrule
\end{tabular}
\caption{Default parameters used in the experiments.}
\label{table:parameters}
\end{table}


\subsection{Standard Whitening vs Whitening Kernels}

In this section, we compare the performance of standard whitening and the whitening kernels in terms of the classification performance of our system (see Figure~\ref{fig:protocol}). Experiments have been conducted by varying the major parameters: \gls{m_expected_time} and the number of filters used in the convolution layer.

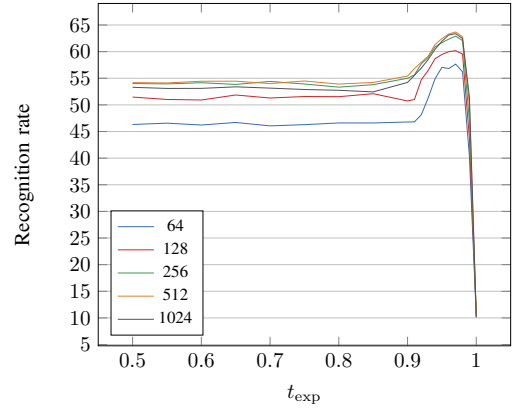
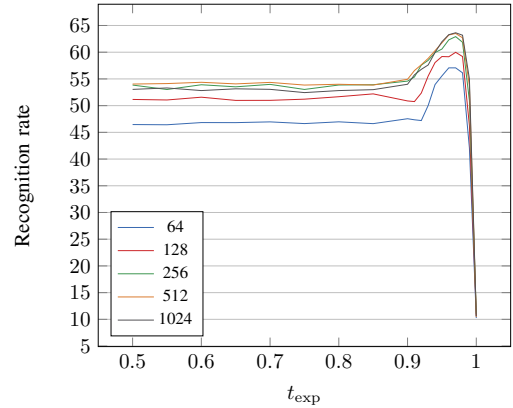
\begin{figure}[ht]
  \centering
  \subfloat[Standard whitening.]{\scalebox{0.8}{\pgfplotstableread[col sep=space]{
0.50 46.32
0.55 46.57
0.60 46.22
0.65 46.7
0.70 46.06
0.75 46.29
0.80 46.61
0.85 46.61
0.90 46.79
0.91 46.81
0.92 48.14
0.93 51.3
0.94 54.9
0.95 57.05
0.96 56.83
0.97 57.66
0.98 56.22
0.99 40.34
1.00 10.12
}\seriea

\pgfplotstableread[col sep=space]{
0.50 51.47
0.55 51.02
0.60 50.92
0.65 51.86
0.70 51.29
0.75 51.57
0.80 51.54
0.85 52.1
0.90 50.74
0.91 51
0.92 54.7
0.93 56.47
0.94 58.66
0.95 59.47
0.96 60.00
0.97 60.18
0.98 59.54
0.99 43.8
1.00 10.56
}\serieb

\pgfplotstableread[col sep=space]{
0.50 54.01
0.55 53.94
0.60 54.16
0.65 53.84
0.70 54.41
0.75 53.92
0.80 53.33
0.85 53.8
0.90 55
0.91 55.64
0.92 57.71
0.93 58.91
0.94 60.9
0.95 61.69
0.96 62.35
0.97 62.92
0.98 62.06
0.99 47.02
1.00 10.5
}\seriec

\pgfplotstableread[col sep=space]{
0.50 54.18
0.55 54.12
0.60 54.44
0.65 54.45
0.70 54
0.75 54.49
0.80 53.91
0.85 54.21
0.90 55.41
0.91 56.82
0.92 58.01
0.93 59.15
0.94 61.34
0.95 62.45
0.96 63.31
0.97 63.69
0.98 62.8
0.99 49.99
1.00 10.29
}\seried

\pgfplotstableread[col sep=space]{
0.50 53.3
0.55 53.09
0.60 53.1
0.65 53.39
0.70 53.15
0.75 52.89
0.80 52.74
0.85 52.44
0.90 54.22
0.91 55.54
0.92 56.91
0.93 58.52
0.94 60.44
0.95 61.95
0.96 63.12
0.97 63.37
0.98 62.46
0.99 51.46
1.00 10.12
}\seriee

\begin{tikzpicture}
    \begin{axis}[
        xlabel=\gls{m_expected_time},
        ylabel=Recognition rate,
		legend pos=south west,
        legend style={font=\fontsize{8}{5}\selectfont},
        ymajorgrids=true,
        ytick distance=5
    ]
    \addplot [color=plotcolor1] plot table {\seriea};
    \addlegendentry{64}
   
    \addplot [color=plotcolor2] plot table {\serieb};
    \addlegendentry{128}

    \addplot [color=plotcolor3] plot table {\seriec};
    \addlegendentry{256}
    
    \addplot [color=plotcolor4] plot table {\seried};
    \addlegendentry{512}

    \addplot [color=plotcolor5] plot table {\seriee};
    \addlegendentry{1024}
    
    \end{axis}
\end{tikzpicture}}\label{sfig:trad_whiten_perf}}\newline
  \centering
  \subfloat[Whitening kernels.]{\scalebox{0.8}{\pgfplotstableread[col sep=space]{
0.50 46.46
0.55 46.42
0.60 46.83
0.65 46.82
0.70 46.96
0.75 46.64
0.80 46.97
0.85 46.63
0.90 47.55
0.91 47.35
0.92 47.21
0.93 50.03
0.94 53.94
0.95 55.56
0.96 57.07
0.97 57.07
0.98 56.14
0.99 42.08
1.00 10.24
}\seriea

\pgfplotstableread[col sep=space]{
0.50 51.16
0.55 51.06
0.60 51.58
0.65 50.98
0.70 50.99
0.75 51.21
0.80 51.67
0.85 52.2
0.90 50.88
0.91 50.76
0.92 52.35
0.93 55.54
0.94 58.06
0.95 59.19
0.96 59.15
0.97 59.98
0.98 59.19
0.99 46.28
1.00 10.68
}\serieb

\pgfplotstableread[col sep=space]{
0.50 53.85
0.55 53.02
0.60 53.91
0.65 53.52
0.70 53.98
0.75 53.03
0.80 53.86
0.85 53.91
0.90 54.59
0.91 55.36
0.92 57.58
0.93 58.47
0.94 59.93
0.95 60.59
0.96 62.31
0.97 62.92
0.98 61.86
0.99 50.93
1.00 11.02
}\seriec

\pgfplotstableread[col sep=space]{
0.50 54.04
0.55 54.13
0.60 54.37
0.65 54.07
0.70 54.35
0.75 53.85
0.80 54.00
0.85 53.85
0.90 54.93
0.91 56.56
0.92 57.65
0.93 58.84
0.94 60.29
0.95 61.74
0.96 63.22
0.97 63.51
0.98 62.70
0.99 53.74
1.00 10.43
}\seried

\pgfplotstableread[col sep=space]{
0.50 53.04
0.55 53.31
0.60 52.82
0.65 53.15
0.70 53.05
0.75 52.44
0.80 52.82
0.85 52.99
0.90 54.00
0.91 55.76
0.92 56.79
0.93 57.61
0.94 59.96
0.95 61.96
0.96 63.28
0.97 63.64
0.98 63.18
0.99 55.01
1.00 10.62
}\seriee

\begin{tikzpicture}
    \begin{axis}[
        xlabel=\gls{m_expected_time},
        ylabel=Recognition rate,
		legend pos=south west,
        legend style={font=\fontsize{8}{5}\selectfont},
        ymajorgrids=true,
        ytick distance=5,
    ]
    \addplot [color=plotcolor1] plot table {\seriea};
    \addlegendentry{64}
   
    \addplot [color=plotcolor2] plot table {\serieb};
    \addlegendentry{128}

    \addplot [color=plotcolor3] plot table {\seriec};
    \addlegendentry{256}
    
    \addplot [color=plotcolor4] plot table {\seried};
    \addlegendentry{512}

    \addplot [color=plotcolor5] plot table {\seriee};
    \addlegendentry{1024}
    
    \end{axis}
\end{tikzpicture}}\label{sfig:patch_whiten_perf}}\newline
  \caption{Classification performance on CIFAR-10 with standard whitening \protect\subref{sfig:trad_whiten_perf} vs whitening kernels \protect\subref{sfig:patch_whiten_perf}.}
  \label{fig:whiten_result}
\end{figure}

Figure~\ref{fig:whiten_result} shows that the behavior of both whitening processes is fairly similar. The reported performances of whitening kernels were obtained using ${9\times 9}$ patches, \gls{m_whiten_coef} = $10^{-2}$ and \gls{m_eigen_ratio} = $1.0$. The performances achieved for each ${t_{exp}}$ and for each number of filters considered are similar. This shows that the approximation of the whitening transformation by convolution kernels performs as well as the original whitening transform. For both methods, \gls{m_expected_time} seems to be optimal around $0.96$.

An in-depth exploration of the parameters of whitening kernels (patch size, whitening coefficient \gls{m_whiten_coef}, and ratio of eigenvectors \gls{m_eigen_ratio}) was conducted. Table~\ref{tab:whiten_coef} shows the results obtained with various kernel sizes, \gls{m_whiten_coef}, and numbers of learned filters (\gls{m_layer_output}). Several observations can be drawn from these results. First, with 64 filters, the performances are overall lower than with 256. However, no strong improvement is observed when using 1024 learned filters. Second, for each configuration using a fixed number of filters, the performances are quite stable regardless of patch size. However, overall, slightly better average performances are obtained when the patch size increases (see $9\times 9$ and $11\times 11$ configurations). A value of $10^{-2}$ for the whitening coefficient \gls{m_whiten_coef} seems more adequate when more filters are used.

\begin{table}
  \centering
  \resizebox{\columnwidth}{!}{
  \begin{tabular}{cccccc}
    \toprule
    \multirow{2}{*}{\textbf{$|\gls{m_layer_output}|$}} & \multirow{2}{*}{\textbf{\gls{m_patch_width} \gls{m_patch_height}}} & \multicolumn{4}{c}{\textbf{\gls{m_whiten_coef}}}\\
    & & \textbf{$10^{-1}$} & \textbf{$10^{-2}$} & \textbf{$10^{-3}$} & \textbf{$10^{-4}$}\\
    \midrule
    \multirow{4}{*}{\textbf{64}} & \textbf{$5\times5$} & 51.36$\pm$0.48 & 56.3$\pm$0.09 & 53.49$\pm$0.09 & 49.89$\pm$0.77 \\
    & \textbf{$7\times7$} & 53.04$\pm$0.31 & 56.74$\pm$0.21 & 54.93$\pm$0.29 & 49.74$\pm$0.88 \\
    & \textbf{$9\times9$} & 53.07$\pm$0.27 & 57.04$\pm$0.10 & \textbf{60.13$\pm$0.23} & 49.96$\pm$1.12 \\
    & \textbf{$11\times11$} & 53.65$\pm$0.14 & 57.04$\pm$0.05 & 55.2$\pm$0.29 & 50.29$\pm$0.44 \\
    \midrule
    \multirow{4}{*}{\textbf{256}} & \textbf{$5\times5$} & 59.32$\pm$0.02 & 62.02$\pm$0.16 & 58.62$\pm$0.3 & 54.46$\pm$0.27 \\
    & \textbf{$7\times7$} & 60.05$\pm$0.26 & 62.55$\pm$0.19 & 58.76$\pm$0.61 & 54.56$\pm$0.27 \\
    & \textbf{$9\times9$} & 59.81$\pm$0.39 & \textbf{63.72$\pm$0.39} & 59.12$\pm$0.34 & 54.62$\pm$0.18 \\
    & \textbf{$11\times11$} & 60.13$\pm$0.29 & 62.83$\pm$0.63 & 58.99$\pm$0.19 & 55.19$\pm$0.56 \\
    \midrule
    \multirow{4}{*}{\textbf{1024}} & \textbf{$5\times5$} & 60.13$\pm$0.37 & 62.91$\pm$0.34 & 57.98$\pm$0.2 & 53.94$\pm$0.13 \\
    & \textbf{$7\times7$} & 60.17$\pm$0.24 & 63.63$\pm$0.51 & 58.47$\pm$0.67 & 54.29$\pm$0.29 \\
    & \textbf{$9\times9$} & 60.72$\pm$0.37 & \textbf{63.78$\pm$0.13} & 58.71$\pm$0.40 & 53.54$\pm$0.42 \\
    & \textbf{$11\times11$} & 61.18$\pm$0.07 & 62.91$\pm$0.34 & 58.87$\pm$0.55 & 53.57$\pm$0.64 \\
    \bottomrule
  \end{tabular}
  }
  \caption{Recognition rate (\%) w.r.t. patch size and \gls{m_whiten_coef} (\gls{m_eigen_ratio} = $1.0$).}
  \label{tab:whiten_coef}
\end{table}

Table~\ref{tab:eigen_ratio} reports the performances with varying patch sizes, \gls{m_eigen_ratio}, and numbers of learned filters. Among all configurations having a given number of filters, the performances are similar. This shows once more the stability and the genericity of the whitening kernels. We can also see that the steep increase in performance from 64 to 256 learned filters is not present when increasing the learning capacity from 256 to 1024 filters. Values of 0.75 and 1.00 for \gls{m_eigen_ratio} provide the best results.

\begin{table}
  \centering
  \resizebox{\columnwidth}{!}{
  \begin{tabular}{cccccc}
    \toprule
    \multirow{2}{*}{\textbf{$|\gls{m_layer_output}|$}} & \multirow{2}{*}{\textbf{\gls{m_patch_width} \gls{m_patch_height}}} & \multicolumn{4}{c}{\textbf{\gls{m_eigen_ratio}}}\\
    & & \textbf{$0.25$} & \textbf{$0.50$} & \textbf{$0.75$} & \textbf{$1.00$}\\
    \midrule
    \multirow{4}{*}{\textbf{64}} & \textbf{$5\times5$} & 50.44$\pm$0.28 & 55.99$\pm$0.11 & 55.85$\pm$0.17 & 56.17$\pm$0.44 \\
    & \textbf{$7\times7$} & 53.32$\pm$0.45 & 56.83$\pm$0.04 & 56.72$\pm$0.17 & 56.60$\pm$0.23 \\
    & \textbf{$9\times9$} & 54.64$\pm$0.06 & 56.88$\pm$0.16 & 57.11$\pm$0.19 & 56.86$\pm$0.03 \\
    & \textbf{$11\times11$} & 55.76$\pm$0.07 & 57.28$\pm$0.23 & 57.12$\pm$0.28 & \textbf{57.33$\pm$0.06} \\
    \midrule
    \multirow{4}{*}{\textbf{256}} & \textbf{$5\times5$} & 57.75$\pm$0.03 & 61.60$\pm$0.04 & 61.53$\pm$0.20 & 61.95$\pm$0.37 \\
    & \textbf{$7\times7$} & 59.43$\pm$0.45 & 62.19$\pm$0.17 & 62.25$\pm$0.14 & 62.41$\pm$0.36 \\
    & \textbf{$9\times9$} & 60.00$\pm$0.07 & 62.37$\pm$0.39 & 62.57$\pm$0.10 & 62.41$\pm$0.21 \\
    & \textbf{$11\times11$} & 61.35$\pm$0.29 & 62.56$\pm$0.91 & \textbf{62.97$\pm$0.28} & 62.70$\pm$0.63 \\
    \midrule
    \multirow{4}{*}{\textbf{1024}} & \textbf{$5\times5$} & 57.9$\pm$0.30 & 62.87$\pm$0.32 & 62.80$\pm$0.35 & 62.88$\pm$0.45 \\
    & \textbf{$7\times7$} & 59.24$\pm$0.44 & 63.4$\pm$0.19 & 63.49$\pm$0.44 & 63.63$\pm$0.28 \\
    & \textbf{$9\times9$} & 59.11$\pm$0.27 & 63.70$\pm$0.09 & 63.39$\pm$0.46 & 63.61$\pm$0.38 \\
    & \textbf{$11\times11$} & 60.71$\pm$0.46 & 62.87$\pm$0.32 & 63.72$\pm$0.42 & \textbf{63.82$\pm$0.45} \\
    \bottomrule
  \end{tabular}
  }
  \caption{Recognition rate (\%) w.r.t. patch size and \gls{m_eigen_ratio} (\gls{m_whiten_coef} = $10^{-2}$).}
  \label{tab:eigen_ratio}
\end{table}

These results show that the benefits brought by whitening kernels are stable and can generalize to a wide set of settings.

\subsection{On-center~/~Off-center Filtering vs Whitening}

\begin{table}
    \centering
    \begin{tabular}{lcc}
        \toprule
        {Method} & {$64$ filters} & {$1,024$ filters}\\
        \midrule
        {On-center~/~off-center (grayscale)~\cite{falez2019unsupervised}} & $45.37\%$ & $52.77\%$ \\
        {On-center~/~off-center (color)~\cite{falez2019unsupervised}} & $48.27\%$ & $56.93\%$ \\
        {Standard whitening} & $\mathbf{57.66\%}$ & $63.37\%$ \\
        {Whitening kernels} & $57.07\%$ & $\mathbf{63.64\%}$ \\
        \bottomrule
    \end{tabular}
    \caption{Performance of whitening versus on-center~/~off-center coding on CIFAR-10.}
    \label{tab:perfs}
\end{table}

\begin{figure}[ht]
\centering
\subfloat[On-center~/~off-center (color) + STDP ($\gls{m_stdp_beta} = 3.0$, $\gls{m_expected_time} = 0.90$).]{
 \begin{tabular}{c}
 \fbox{\includegraphics[width=0.11\columnwidth]{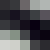}}
 \fbox{\includegraphics[width=0.11\columnwidth]{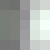}}
 \fbox{\includegraphics[width=0.11\columnwidth]{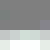}}
 \fbox{\includegraphics[width=0.11\columnwidth]{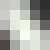}}
 \fbox{\includegraphics[width=0.11\columnwidth]{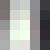}}\\
 \fbox{\includegraphics[width=0.11\columnwidth]{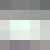}}
 \fbox{\includegraphics[width=0.11\columnwidth]{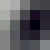}}
 \fbox{\includegraphics[width=0.11\columnwidth]{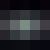}}
 \fbox{\includegraphics[width=0.11\columnwidth]{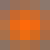}}
 \fbox{\includegraphics[width=0.11\columnwidth]{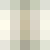}}
 \end{tabular}
}\newline
 \subfloat[Standard whitening + STDP ($\gls{m_stdp_beta} = 3.0$, $\gls{m_expected_time} = 0.97$).]{
  \begin{tabular}{c}
 \fbox{\includegraphics[width=0.11\columnwidth]{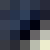}}
 \fbox{\includegraphics[width=0.11\columnwidth]{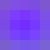}}
 \fbox{\includegraphics[width=0.11\columnwidth]{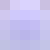}}
 \fbox{\includegraphics[width=0.11\columnwidth]{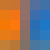}}
 \fbox{\includegraphics[width=0.11\columnwidth]{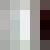}}\\
 \fbox{\includegraphics[width=0.11\columnwidth]{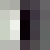}}
 \fbox{\includegraphics[width=0.11\columnwidth]{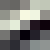}}
 \fbox{\includegraphics[width=0.11\columnwidth]{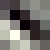}}
 \fbox{\includegraphics[width=0.11\columnwidth]{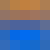}}
 \fbox{\includegraphics[width=0.11\columnwidth]{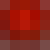}}\\
 \end{tabular}
}\newline
 \subfloat[Kernel whitening + STDP ($\gls{m_stdp_beta} = 3.0$, $\gls{m_expected_time} = 0.97$).]{
  \begin{tabular}{c}
 \fbox{\includegraphics[width=0.11\columnwidth]{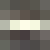}}
 \fbox{\includegraphics[width=0.11\columnwidth]{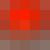}}
 \fbox{\includegraphics[width=0.11\columnwidth]{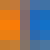}}
 \fbox{\includegraphics[width=0.11\columnwidth]{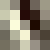}}
 \fbox{\includegraphics[width=0.11\columnwidth]{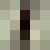}}\\
 \fbox{\includegraphics[width=0.11\columnwidth]{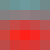}}
 \fbox{\includegraphics[width=0.11\columnwidth]{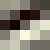}}
 \fbox{\includegraphics[width=0.11\columnwidth]{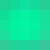}}
 \fbox{\includegraphics[width=0.11\columnwidth]{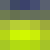}}
 \fbox{\includegraphics[width=0.11\columnwidth]{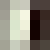}}\\
 \end{tabular}
 }\newline
  \subfloat[Standard whitening + autoencoders (images taken from~\cite{coates2011analysis}).]{
  \hspace*{-.5cm}
  \begin{tabular}{c}
 \fbox{\includegraphics[width=0.11\columnwidth]{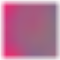}}
 \fbox{\includegraphics[width=0.11\columnwidth]{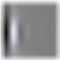}}
 \fbox{\includegraphics[width=0.11\columnwidth]{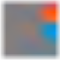}}
 \fbox{\includegraphics[width=0.11\columnwidth]{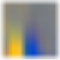}}
 \fbox{\includegraphics[width=0.11\columnwidth]{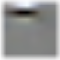}}\\
 \fbox{\includegraphics[width=0.11\columnwidth]{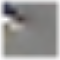}}
 \fbox{\includegraphics[width=0.11\columnwidth]{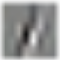}}
 \fbox{\includegraphics[width=0.11\columnwidth]{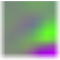}}
 \fbox{\includegraphics[width=0.11\columnwidth]{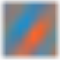}}
 \fbox{\includegraphics[width=0.11\columnwidth]{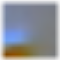}}
 \end{tabular}
 }
 \caption{Samples of filters learned on CIFAR-10 with different pre-processing and learning methods.}
 \label{fig:filter_samples}
\end{figure}

Table~\ref{tab:perfs} compares the performances on CIFAR-10 of whitening to the baseline on-center~/~off-center coding, for $|\gls{m_layer_output}| = 64$ features and $|\gls{m_layer_output}| = 1024$ features. Whitening provides much better results than color on-center~/~off-center coding: +18\% (+9 percentage points) with 64 filters and +11\% (+6 pp.) with 1,024 filters. This may be due to its ability to retain color information and all spatial frequencies, whereas on-center~/~off-center coding only encodes edge information and a limited range of spatial frequencies.

Figure~\ref{fig:filter_samples} shows samples of features learned on CIFAR-10 with STDP with the three pre-processing approaches, as well as features learned by an auto-encoder with standard whitening. Whereas the filters learned with on-center~/~off-center coding (Figure~\ref{fig:filter_samples}(a)) are almost only oriented edges, filters learned with whitening (Figures~\ref{fig:filter_samples}(b) and \ref{fig:filter_samples}(c)) include both oriented edges and oriented color patterns. These filters are much closer to the ones that can be learned on whitened data by an auto-encoder (Figure~\ref{fig:filter_samples}(d), taken from~\cite{coates2011analysis}), but also the ones learned by DNNs on non-whitened data (see for instance~\cite{krizhevsky12imagenet}). The main difference with auto-encoder features is that they are more localized, which may be due to our filters being smaller in size ($5\times 5$ pixels vs $8\times 8$ pixels in~\cite{coates2011analysis}).

\subsection{Cross-dataset experiments}

Since computing whitening transformations is computationally expensive and not suited to neuromorphic hardware, the learned transformations should be able to be reused on different datasets, to avoid re-training them. In order to test this ability, whitening kernels are computed independently from CIFAR-10 and STL-10, respectively. Then, the CIFAR-10 dataset is pre-processed using whitening kernels learned on STL-10, and reversely. Each whitened dataset is then fed into our classification system and its accuracy is measured. 

Table~\ref{tab:cross_dataset} shows the result obtained using different configurations. We used several numbers of filters. We fixed the following parameters : (a) \gls{m_eigen_ratio}=1; (b) \gls{m_whiten_coef} = $10^{-2}$, and (c) patch size = $9\times9$. We report the results obtained either using whitening kernels computed on the same dataset or whitening kernels computed on the other dataset. Regardless of the underlying configuration, the difference of the recognition rates between whitening kernels trained on same dataset and trained on a different datasets is negligible. In almost all cases, the difference between the two configurations is close to zero or statistically not significant (smaller than the standard deviation). Thus, whitening kernels are dataset independent and can be computed once, an reused on multiple datasets.

\begin{table}
  \centering
  \resizebox{\columnwidth}{!}{
  \begin{tabular}{l|ccc|ccc}
    \toprule
    \multirow{2}{*}{$|\gls{m_layer_output}|$}& \multicolumn{3}{c|}{\textbf{CIFAR-10}} & \multicolumn{3}{c}{\textbf{STL-10}}\\
    & \textbf{CIFAR-10} & \textbf{SLT-10} & $\Delta$ & \textbf{STL-10} & \textbf{CIFAR-10} & $\Delta$\\
    \midrule
      \textbf{64} & 57.66$\pm$0.44 & 57.09$\pm$0.11 & -0.57 & 57.08$\pm$0.44 & 56.97$\pm$0.34 & +0.11 \\
      \textbf{128} & 60.18$\pm$0.29 & 59.95$\pm$0.08 & -0.23 & 58.93$\pm$0.25 & 58.74$\pm$0.48 & +0.19 \\
      \textbf{256} & 62.92$\pm$0.10 & 62.77$\pm$0.30 & -0.15 & 59.86$\pm$0.32 & 59.74$\pm$0.43 & +0.12\\
      \textbf{512} & 63.69$\pm$0.16 & 63.78$\pm$0.18 & +0.09 & 60.73$\pm$0.46 & 60.72$\pm$0.42 & +0.01 \\
      \textbf{1024} & 63.37$\pm$0.21 & 63.80$\pm$0.56 & +0.43 & 63.29$\pm$0.11 & 62.94$\pm$0.42 & -0.33 \\
    \bottomrule
  \end{tabular}
  }
  \caption{Performances obtained in a cross-dataset configuration (first header row: dataset used for classification, second header row: dataset used to generate the whitening kernels.}
  \label{tab:cross_dataset}
\end{table}

\section{Conclusion}

SNNs trained with \Gls{stdp} are good candidates to produce ultra-low power neural networks. However, their performances is currently far behind \glspl{dnn}. Notably, \gls{stdp} cannot learn effective features on real-word color images. On-center~/~off-center coding, used to pre-process images in this context, is partially responsible for it, as it filters only a subrange of spatial frequencies from the original images. In this paper, we showed that pre-processing images with whitening allows to learn more effective features, closer to the ones learned with standard neural networks. Implementing whitening on neuromorphic hardware may not be trivial, so we also propose to approximate whitening with convolution kernels to facilitate its implementation. It yields almost the same performance as traditional whitening. Cross-dataset experiments show stable performance of the whitening kernels over datasets, making it possible to learn a single set of kernels to process different datasets, making it even more suitable in a low-power context.


\bibliographystyle{plain}
\bibliography{main}

\end{document}